\documentclass[11pt]{article}

\usepackage[final]{acl}

\usepackage{times}
\usepackage{latexsym}

\usepackage[T1]{fontenc}

\usepackage[utf8]{inputenc}

\usepackage{microtype}

\usepackage{inconsolata}

\usepackage{graphicx}
\usepackage[most]{tcolorbox}
\usepackage{xcolor}
\usepackage{xspace}
\usepackage{cuted}
\usepackage{marvosym}
\usepackage{booktabs}
\usepackage{amsfonts}
\usepackage{arydshln}

\newcommand{\ourmethod}{MotionInsight\xspace}
\newcommand{\ourdataset}{VidMotion\xspace}
\newcommand{\myparagraph}[1]{\vspace{0.1cm}\noindent\textbf{#1}}

\title{\ourmethod: Diagnosing Object Motion Deficiencies in Generated Videos}

\author{
 \textbf{Jiahao Zhan\textsuperscript{1,2}},
 \textbf{Yongrui Ma\textsuperscript{1,2}},
 \textbf{Qunliang Xing\textsuperscript{2}},
 \textbf{Xuanyu Zhang\textsuperscript{4}},
\\
 \textbf{Jingqi Tong\textsuperscript{3}},
 \textbf{Junlin Li\textsuperscript{2}},
 \textbf{Li Zhang\textsuperscript{2}},
 \textbf{Shijie Zhao\textsuperscript{2,$\dagger$,\Letter}},
  \textbf{Tianfan Xue\textsuperscript{1,5,\Letter}}
\\
\\
 \textsuperscript{1}MMLab, CUHK,
 \textsuperscript{2}ByteDance Inc.,
 \textsuperscript{3}Fudan University,
 \textsuperscript{4}Peking University,
 \textsuperscript{5}CPII under InnoHK
 \\
\small{
   \textsuperscript{$\dagger$}Project Lead
   \quad
   \Letter\ Correspondence:
   \href{mailto:zhaoshijie.0526@bytedance.com}{zhaoshijie.0526@bytedance.com},
   \href{mailto:tfxue@ie.cuhk.edu.hk}{tfxue@ie.cuhk.edu.hk}
}
}

\begin{document}
\maketitle
\begin{strip}
\begin{center}
\vspace{-2cm}

\begin{minipage}{\textwidth}
    \centering
    \includegraphics[width=\textwidth]{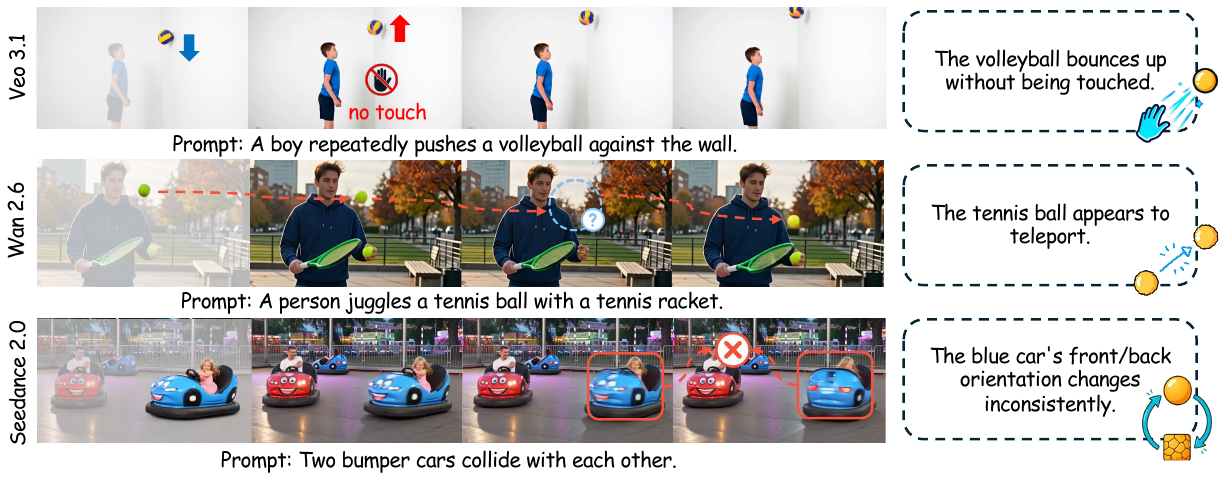}
    \vspace{-0.5cm}
    \captionsetup{type=figure,width=\textwidth}
    
\captionof{figure}{
Examples of motion deficiencies in generated videos. The left-side sampled frames highlight the object exhibiting the motion failure in each video, while the right-side panels describe the corresponding failures.
}
\vspace{-0.5cm}

    \label{fig:teaser}
\end{minipage}

\end{center}

\end{strip}

\begin{abstract}
Despite rapid progress in video generation models, they still exhibit obvious motion deficiencies, often manifested as incorrect object motion. 
However, most existing video quality evaluations focus on aesthetic quality or text-video alignment. 
To address this gap, we study object-centric motion fidelity assessment, evaluating target objects along object consistency, motion continuity, and physical plausibility.
To achieve this, we first introduce \ourdataset, a diagnostic dataset of 6,879 videos with designated moving objects and fine-grained annotations including dimension-wise scores and failure causes.
We further propose \ourmethod, a diagnostic evaluator that shifts assessment from implicit RGB-frame observation to explicit motion-space diagnosis.
By constructing motion-aware representations, \ourmethod makes subtle motion deficiencies more observable.
We also introduce motion-specific rewards during GRPO to transform observed motion into a diagnostic assessment. 
Experiments demonstrate that \ourmethod provides an effective basis for diagnosing object motion deficiencies, producing human-aligned scores along three dimensions and grounded explanations. The code is publicly available at \url{https://github.com/JohnZhan2023/MotionInsight}.

\end{abstract}

\section{Introduction}

Video generation has undergone rapid development in visual quality~\cite{Cogvideo}, while it may still generate unrealistic motion in dynamic scenarios~\cite{bansal2024videophy}. 
Current video evaluation methods~\cite{he2025videoscore2} emphasize aesthetic quality and text-video alignment, while paying limited attention to whether objects move in a stable, continuous, and physically reasonable manner. 
As shown in Figure~\ref{fig:teaser}, visually compelling videos may still contain severe motion deficiencies. 
Such deficiencies restrict the applications in film production~\cite{jiang2025vace} and world modeling~\cite{zhan2026perpetualwonder}, highlighting the need for a dedicated evaluation task that systematically diagnoses motion deficiencies in generated videos.

To this end, we formulate a diagnostic task of object-centric motion fidelity assessment. 
We focus the evaluation on designated moving objects, including people, animals, vehicles, and everyday physical objects, since they often attract substantial visual attention in videos.
This object-centric formulation provides a fine-grained basis for diagnosis by localizing the assessment to a concrete target. 
We further assess object motion fidelity along three complementary dimensions: object consistency, motion continuity, and physical plausibility, whose definitions are shown in Table~\ref{tab:motion_dimensions}.
Beyond predicting dimension-wise scores, a diagnostic assessment should provide grounded explanations of concrete motion failures, such as the volleyball in the top row of Figure~\ref{fig:teaser} bouncing upward without any physical contact.

To support this study, we build \ourdataset, a diagnostic dataset for object-centric motion fidelity assessment. 
Most prior video evaluation datasets primarily provide holistic quality scores, which offer limited interpretability and cannot support fine-grained diagnosis of object motion deficiencies.
In contrast, each video in \ourdataset designates a moving object to be evaluated and provides diagnostic annotations that include scores along three dimensions and failure causes when object motion deficiencies occur.
In total, \ourdataset contains 5,713 generated videos from nine representative models and 1,166 real videos, covering most everyday dynamic scenarios.
Based on the human annotations, we further reveal a substantial gap between generated and real motion, as shown in Table~\ref{tab:model_comparison}. 
This gap raises a natural question: Can existing evaluators diagnose object motion deficiencies in generated videos?

Unfortunately, as shown in Table~\ref{tab:main_comparison}, existing evaluators remain limited in identifying subtle motion deficiencies. 
Although VLMs have strong semantic understanding, they typically rely on sampled RGB frames to infer motion.
Such RGB-frame inputs provide only partial temporal observations and contain substantial background redundancy.
As a result, subtle motion deficiencies, such as fingers becoming distorted, can be easily diluted by redundant visual content and become difficult to capture.
This becomes especially problematic for human-aligned evaluation, since human judgments are often driven by the most severe local artifacts~\cite{human_visual}, while VLM-based evaluators may miss them.
Such missed evidence further limits their ability to provide grounded explanations.

To address these challenges, we propose \ourmethod, which diagnoses object motion deficiencies in an explicit motion space.
Specifically, \ourmethod constructs motion-aware representations from object tracking features and global camera motion, rather than solely relying on sampled RGB frames.
Such structured motion evidence makes the evaluator more sensitive to subtle deficiencies that may be diluted in RGB-frame observations.
Moreover, to help the VLM interpret this motion evidence, we further perform motion description alignment, which uses automatically generated motion descriptions to align the encoded motion embeddings with the VLM's semantic space.
Finally, to turn motion evidence into human-aligned diagnosis, we introduce motion-specific rewards for Group Relative Policy Optimization~(GRPO)~\cite{guo2025deepseek}, derived from multi-dimensional scores and failure-cause supervision.
These rewards align the evaluator's scoring criteria with human judgments while enabling diagnostic insights into concrete object motion failures.

Experiments show that \ourmethod aligns well with human judgments, producing accurate dimension-wise scores, grounded diagnostic reasoning, and strong sensitivity to localized motion failures.

\begin{table}[t]
\centering
\small
\setlength{\tabcolsep}{0pt}
\renewcommand{\arraystretch}{1.08}
\begin{tabular}{p{0.10\linewidth}@{\hspace{6pt}}p{0.82\linewidth}}
\toprule
\textbf{Dim.} & \textbf{Description} \\
\midrule
OC & Stable identity, appearance, and structure during motion. \\
MC & Smooth and continuous trajectory without abrupt jumps or jitter. \\
PP & Motion consistent with forces and basic physical dynamics. \\
\bottomrule
\end{tabular}
\caption{Definitions of the three dimensions in object-centric motion fidelity assessment: object consistency (OC), motion continuity (MC), and physical plausibility (PP).}
\label{tab:motion_dimensions}
\end{table}

\begin{table*}[t]
\centering
\small
\setlength{\tabcolsep}{5pt}
\begin{tabular}{lccc}
\toprule
\textbf{Model} & \textbf{Object Consistency}$\uparrow$ & \textbf{Motion Continuity}$\uparrow$ & \textbf{Physical Plausibility}$\uparrow$ \\
\midrule
\multicolumn{4}{l}{\textit{Open-source models}} \\
Wan 2.2~\cite{Wan} & $3.420 \pm 0.441$& $3.634 \pm 0.530$& $2.355 \pm 0.261$\\
LTX 2.1~\cite{LTXVideo} & $1.978 \pm 0.690$& $3.512 \pm 0.546$& $2.930 \pm 0.402$\\
LongCat-Video~\cite{team2025longcat} & $1.335 \pm 0.214$& $2.242 \pm 0.699$& $1.821 \pm 0.582$\\
Cosmos-Predict 2.5~\cite{ali2025world} & $2.088 \pm 0.649$& $3.109\pm 0.679$& $2.156 \pm 0.472$\\
HunyuanVideo-1.5~\cite{hunyuanvideo} & $3.223 \pm 0.504$& $3.939 \pm 0.672$& $2.370 \pm 0.274$\\
\midrule
\multicolumn{4}{l}{\textit{Proprietary models}} \\
Wan 2.6~\cite{Wan} & $3.577 \pm 0.426$& $3.746 \pm 0.644$& $2.461 \pm 0.314$\\
Sora 2~\cite{sora} & $3.792 \pm 0.375$& $4.113 \pm 0.462$& $3.507 \pm 0.270$\\
Veo 3.1~\cite{veo3} & $3.189 \pm 0.532$& $3.636 \pm 0.698$& $2.250 \pm 0.382$\\
Seedance 2.0~\cite{seedance2025seedance} & $3.810 \pm 0.379$& $4.143 \pm 0.573$& $3.081 \pm 0.250$\\
\midrule
\textit{Real Video} & $4.740 \pm 0.269$& $4.915 \pm 0.265$& $4.814 \pm 0.112$\\
\bottomrule
\end{tabular}
\caption{Comparison of video generation models and real videos on object consistency, motion continuity, and physical plausibility, evaluated on the 166 challenging prompts in \ourdataset-Test. Values are reported as mean $\pm$ standard deviation.}
\label{tab:model_comparison}
\end{table*}

\section{Related Work}
\myparagraph{Video motion evaluation and benchmarks.}
Existing benchmarks for video generation mainly focus on overall visual quality and text-to-video alignment, such as VBench~\cite{vbench}, VideoScore2~\cite{he2025videoscore2}, and T2V-CompBench~\cite{sun2025t2v}. 
To address the evaluation of motion, several studies incorporate low-level or structured signals. 
For example, VBench uses optical-flow-based metrics, but optical flow may fail to maintain reliable temporal correspondence when the target object suddenly disappears. 
VMBench~\cite{ling2025vmbench} evaluates motion consistency via rule-based filtering of tracks, while HumanScore~\cite{fang2026humanscore} leverages 3D human body models to assess human motion realism. 
Similarly, WorldScore~\cite{worldscore} introduces Structure-from-Motion~\cite{sfm} to measure whether generated videos follow physical camera constraints. These approaches rely heavily on priors for specific tasks, which makes them inherently constrained to narrow domains and difficult to generalize to broader video generation scenarios. Other studies evaluate physical plausibility or world modeling capabilities of video generators~\cite{meng2024towards, hu2025benchmarking}, often relying on predefined rules or constrained scenarios. 
As a result, they cannot serve as general evaluators for diagnosing object motion deficiencies.

\myparagraph{VLM-based evaluators for video generation.} Recent works have explored VLM-based evaluators for video generation~\cite{qin2024worldsimbench, bansal2024videophy, zhang2026vq, zhao2025reasoning, wang2025unified, wu2022fast}, leveraging the strong generalization ability of VLMs. 
Some recent works further improve video perception by strengthening temporal attention~\cite{motamed2025travl} or reducing redundant visual tokens~\cite{shi2026attend}.
Nevertheless, these methods still rely solely on sampled RGB frames to perceive motion. Many object motion deficiencies manifest as subtle pixel-level changes in RGB space and can be easily diluted during visual-token aggregation.
In contrast, \ourmethod models the object motion in motion space, enabling greater sensitivity to local artifacts.

\section{\ourdataset Dataset}

\ourdataset is a diagnostic dataset designed for object-centric motion fidelity assessment.
Unlike prior evaluation datasets that mainly provide overall scores for physical plausibility, artifact severity, or video realism~\cite{bansal2025videophy, zhang2026physion}, \ourdataset evaluates the motion fidelity of a designated object.
Each video is associated with a designated moving object and annotated with three-dimensional motion fidelity scores and fine-grained failure causes.
Together, these annotations provide a strong basis for diagnosing object motion deficiencies by localizing to a concrete target, explaining the failure cause, and measuring its severity along different motion dimensions.
Since our focus is perceptual motion fidelity rather than solver-based correctness, the annotations aim to capture whether the target object's motion appears realistic from human experience. 
So real videos are evaluated under the same human annotation protocol rather than assigned perfect scores by default, enabling direct comparison between generated and natural object motion.
In total, \ourdataset contains 6,879 annotated videos.
Figure~\ref{fig:dataset} summarizes the overall data construction and annotation pipeline.

\subsection{Dataset Construction}
To build \ourdataset, we start from real-world videos collected from open-source datasets~\cite{caelles20192019, wu2016physics, chow2025physbench, kay2017kinetics, huang2019got, motamed2025travl} and additional curated sources, prioritizing samples in which one salient moving object dominates the motion. After screening by five human experts, we obtain a final set of 1,166 real videos.

We then derive textual prompts and targets for these videos through captioning with Gemini 3.1 Pro~\cite{team2024gemini}, followed by human verification. Using the resulting prompts, we generate corresponding videos with a diverse set of video generation models, including open-source models such as Wan 2.2~\cite{Wan}, LTX 2.1~\cite{LTXVideo}, LongCat-Video~\cite{team2025longcat}, Cosmos-Predict 2.5~\cite{ali2025world}, and HunyuanVideo-1.5~\cite{hunyuanvideo}, as well as proprietary models including Wan 2.6~\cite{Wan}, Sora 2~\cite{sora}, Veo 3.1~\cite{veo3}, and Seedance 2.0~\cite{seedance2025seedance}. For proprietary models, we generate videos only for a challenging subset of 166 prompts manually selected from the 1,166 prompts. Since videos from these proprietary models are not included in \ourdataset-Train, this design makes \ourdataset-Test more challenging and allows us to evaluate generalization to unseen generators.

In this way, each prompt is paired with a corresponding real video and multiple generated videos from different models. After filtering out invalid samples with severe degeneration or unusable content, we use the 1,386 videos associated with the 166 challenging prompts as \ourdataset-Test, which serves as our benchmark split. The remaining 5,493 videos are used as \ourdataset-Train.

\subsection{Human Annotation}
We collect human annotations on the motion fidelity of the designated object along three dimensions: object consistency, motion continuity, and physical plausibility. 
For videos judged to contain motion deficiencies, annotators further select one or more applicable failure causes from 12 predefined candidates, which provide diagnostic explanations for the underlying motion artifacts.

In total, 21 annotators participated in the annotation process. Before annotation, all annotators underwent training, completed a trial annotation assessment, and reviewed representative examples with reference labels. Each video is independently evaluated by three annotators, each of whom provides three-dimensional scores, a confidence level, and applicable failure causes. Krippendorff's $\alpha$~\cite{krippendorff2011computing} averaged over the three dimensions reaches 0.7420, indicating reliable annotations. More annotation details are provided in Appendix~\ref{data_annotation}. 
We aggregate the three annotators' scores using the Mean Opinion Score (MOS) to obtain the final three-dimensional motion fidelity scores, and take the intersection of their selected failure causes as the final multi-label failure-cause annotations to ensure label reliability. 
Additional analysis of \ourdataset is included in Appendix~\ref{dataset_analysis}.

\section{\ourmethod}

\begin{figure*}[t]
    \centering
    \includegraphics[width=\linewidth]{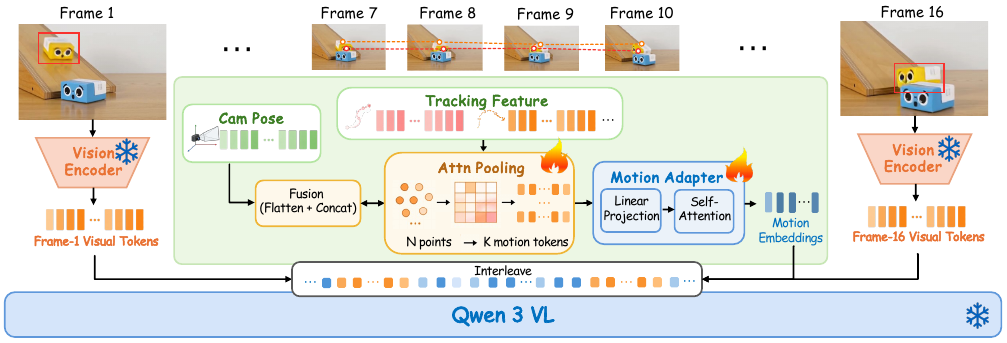}
    \caption{Overview of \ourmethod. Given an input video and a target prompt, we uniformly sample RGB frames as the standard visual input. Then, we extract motion features from the frames using ViPE and CoTracker3, and aggregate them into a motion-aware representation, which is fed into the VLM. }
    \label{fig:method}
\end{figure*}

Our goal is to assess the motion fidelity of a designated object in a video.
Given a video $\mathbf{v}=\{f_t\}_{t=1}^{T}$ and a target object prompt $o$, our evaluator $\mathcal{E}$ outputs a three-dimensional motion score vector and corresponding diagnostic reasoning, $(\mathbf{s}, \mathbf{r}) = \mathcal{E}(\mathbf{v}, o)$. 
Here, $\mathbf{s}$ covers object consistency, motion continuity, and physical plausibility.
For VLM-based evaluators, $\mathbf{r}$ corresponds to the explicit reasoning text generated before the final scores to justify the scoring results.
In the benchmark setting, the target object prompt $o$ is provided by the dataset annotation or specified by the user.
When no target object is given, such as in reward-model applications for video generation, we first prompt a VLM to identify salient moving objects in the video, apply \ourmethod to each identified object, and aggregate the resulting object-centric scores into a video-level signal, as described in Appendix~\ref{rm_application}.

Since RGB-frame observations provide only sparse temporal information and contain substantial background redundancy, subtle motion deficiencies can be difficult to perceive. 
Our key design principle is therefore to explicitly represent the complete motion in a structured motion space. 
Based on this motion evidence, \ourmethod performs human-aligned diagnosis through semantic alignment with the VLM and preference alignment with human judgments.
As illustrated in Figure~\ref{fig:method}, \ourmethod complements sampled frames with motion-aware representations, making object motion deficiencies observable~(Section~\ref{motion_representation}).
Next, we align these motion embeddings with the VLM's semantic space using automatically generated motion descriptions~(Section~\ref{modality_alignment}). 
Finally, we design motion-specific rewards for GRPO to align the resulting diagnosis with human judgments~(Section~\ref{grpo}).

\subsection{Motion-Aware Representations}
\label{motion_representation}

Motion-aware representations aim to make object motion deficiencies explicit beyond RGB-frame observations.
However, the motion in a video is entangled with both object motion and camera motion.
We therefore fuse object motion with camera poses to form motion embeddings, enabling a disentangled understanding of object dynamics.

Specifically, given a video $\mathbf{v}=\{f_t\}_{t=1}^{T}$ and a text prompt $o$ specifying the target object, we first use SAM3~\cite{carion2025sam} to obtain an initialization mask for the target object:
\begin{equation}
m = \mathrm{SAM3}(\mathbf{v}, o).
\end{equation}
Using this mask, we sample query points $\mathcal{P}$ and track them throughout the video using CoTracker3~\cite{karaev2025cotracker3}. This yields frame-wise point-level tracking features $\mathbf{X}=\{\mathbf{x}_t\}_{t=1}^{T}$, where each $\mathbf{x}_t \in \mathbb{R}^{N \times d_o}$ and $N$ is the number of sampled tracked points.

Since the number of tracked points $N$ varies with the size of $m$, we use an attention pooling module with $K$ learnable queries to aggregate the point-level features into a fixed number of object-motion tokens:
\begin{equation}
\mathbf{H}_t = \mathrm{AttnPool}(\mathbf{x}_t) \in \mathbb{R}^{K \times d_m}.
\end{equation}

In parallel, we feed the entire video into ViPE~\cite{huang2025vipe} to estimate the camera poses for all frames:
\begin{equation}
\mathbf{C}=\{\mathbf{c}_t\}_{t=1}^{T}=\mathrm{ViPE}(\mathbf{v}),
\end{equation}
where each $\mathbf{c}_t = [r_t ; \boldsymbol{\tau}_t]$ consists of a 6D rotation parameter $r_t \in \mathbb{R}^{6}$ and a 3D translation parameter $\boldsymbol{\tau}_t \in \mathbb{R}^{3}$ for frame $f_t$. These camera poses provide a global motion reference that helps separate object motion from viewpoint changes.

For each frame, we construct a motion representation by flattening the $K$ object-motion tokens and concatenating them with the corresponding camera poses:
\begin{equation}
\mathbf{z}_t=[\mathrm{vec}(\mathbf{H}_t); \mathbf{c}_t] \in \mathbb{R}^{K d_m + 9}.
\end{equation}
The resulting sequence $\mathbf{Z}=\{\mathbf{z}_t\}_{t=1}^{T}$ is processed by a lightweight Motion Adapter, consisting of a linear projection and a self-attention layer, to produce frame-wise motion embeddings.

As shown in Figure~\ref{fig:method}, we uniformly sample frames from the video and interleave them with the corresponding motion embeddings produced by the Motion Adapter. This interleaved sequence enables the VLM to jointly reason over appearance information and structured motion representations.

\subsection{Motion Description Alignment}
\label{modality_alignment}

Although the motion-aware representations encode object motion and camera poses, they are not directly interpretable by the frozen VLM. We therefore align the extracted tracking features and camera poses with the VLM's semantic space using automatically generated motion descriptions as supervision. 
To obtain supervision without additional human annotations, we design 16 questions about the target's motion. Each video is divided into clips of 16 consecutive frames, which are fed sequentially to the VLM~\cite{bai2025qwen3} to generate clip-level descriptions. These descriptions are then summarized into a video-level motion description.

Following this procedure, we collect 80,721 question-answer pairs from OpenVid~\cite{nan2024openvid} for semantic alignment. We then fine-tune the model with this motion-description supervision for semantic alignment. During this stage, we freeze the VLM backbone and optimize only the motion encoding components, including the attention pooling module and the Motion Adapter.

\subsection{GRPO for Human Preference Alignment}
\label{grpo}

We adopt GRPO~\cite{guo2025deepseek} with two motion-specific rewards: a multi-dimensional reward and a failure-cause reward.
The multi-dimensional reward aligns \ourmethod with human judgments by encouraging accurate prediction of annotated scores along object consistency, motion continuity, and physical plausibility.
The failure-cause reward further refines its diagnostic reasoning by encouraging the evaluator to associate observed motion deficiencies with interpretable failure causes.
Built on the richer and more explicit motion evidence provided by motion-aware representations,
the motion-specific rewards guide \ourmethod to learn human-aligned scoring and grounded reasoning.
Together, these two capabilities constitute the core of a diagnostic evaluator for object-centric motion fidelity.

\myparagraph{Multi-dimensional scoring reward.} 
For the scoring reward, we supervise the model using the three-dimensional scores in \ourdataset-Train. 
Given a query $q$, the model predicts scores for $M$ evaluation dimensions $\{\hat{s}_j\}_{j=1}^{M}$. 
To discourage overconfident scoring, we use a normalized asymmetric error penalty that penalizes overestimation more heavily than underestimation:
\begin{equation}
r^{\mathrm{score}} = 1 - \sum_{j=1}^{M} \lambda_j \, \mathrm{clip}(d_j^2, 0, 1),
\end{equation}
where
\begin{equation}
d_j =
\begin{cases}
\alpha \, \dfrac{|\hat{s}_j - s_j^{\mathrm{gt}}|}{4}, & \text{if } \hat{s}_j > s_j^{\mathrm{gt}}, \\[6pt]
\dfrac{|\hat{s}_j - s_j^{\mathrm{gt}}|}{4}, & \text{otherwise}.
\end{cases}
\end{equation}
Here, $s_j^{\mathrm{gt}}$ denotes the ground-truth MOS for the $j$-th dimension, $\lambda_j$ is its weight, and $\alpha > 1$ controls the additional penalty for overestimation.

\myparagraph{Failure-cause reward.}
To improve \ourmethod's ability to identify motion deficiencies and provide grounded reasoning, we introduce a failure-cause reward using the failure-cause labels from \ourdataset-Train. Given four candidate failure descriptions, the model selects the one that best explains the observed issue. We assign a reward of 1 if the selected option matches the annotated failure causes and 0 otherwise.

\section{Experiments}
\begin{figure*}[t]
    \centering
    \includegraphics[width=\linewidth]{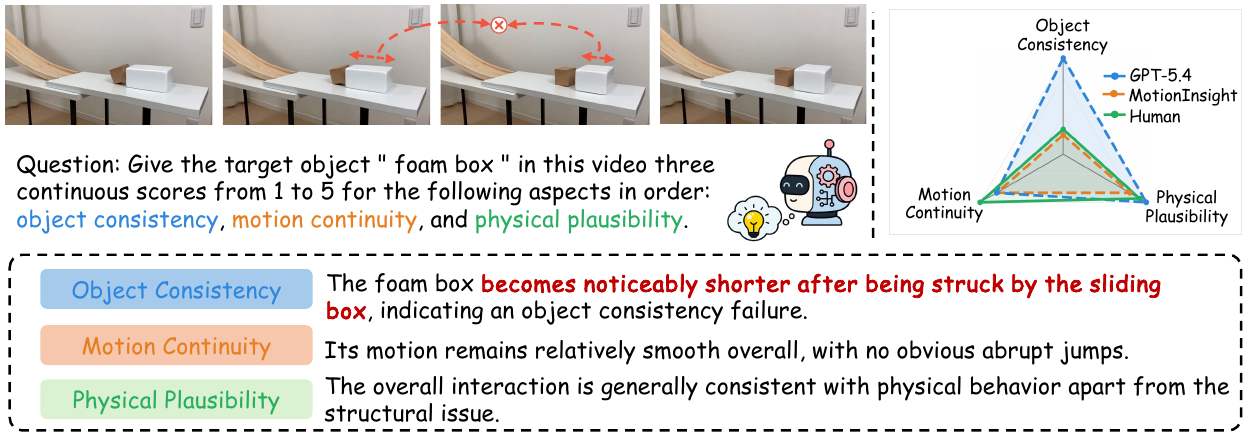}
    \caption{Qualitative result of \ourmethod. The reasoning texts are presented below the video frames, and the scores for the three dimensions are visualized in the radar chart at the top right.}

    \label{fig:qualitative_result}
\end{figure*}

\begin{table*}[t]
\centering
\small
\setlength{\tabcolsep}{4.5pt}
\renewcommand{\arraystretch}{1.02}
\begin{tabular}{@{}lccccccccc@{}}
\toprule
& \multicolumn{3}{c}{\textbf{Object Consistency}} 
& \multicolumn{3}{c}{\textbf{Motion Continuity}} 
& \multicolumn{3}{c}{\textbf{Physical Plausibility}} \\
\cmidrule(lr){2-4} \cmidrule(lr){5-7} \cmidrule(lr){8-10}
\textbf{Method} & \textbf{SRCC}$\uparrow$ & \textbf{PLCC}$\uparrow$ & \textbf{KRCC}$\uparrow$ & \textbf{SRCC}$\uparrow$ & \textbf{PLCC}$\uparrow$ & \textbf{KRCC}$\uparrow$ & \textbf{SRCC}$\uparrow$ & \textbf{PLCC}$\uparrow$ & \textbf{KRCC}$\uparrow$ \\
\midrule
Independent Human & 0.732 & 0.774 & 0.671 & 0.614 & 0.684 & 0.580 & 0.766 & 0.804 & 0.692 \\
\hdashline
GPT-5.4 & 0.370 & 0.400 & 0.322 & 0.276 & 0.303 & 0.243 & 0.246 & 0.257 & 0.203 \\
Gemini 3.1 Pro & 0.333 & 0.318 & 0.281 & 0.218 & 0.217 & 0.193 & 0.166 & 0.162 & 0.144 \\
Qwen-3-VL-8B & 0.074 & 0.092 & 0.065 & 0.055 & 0.062 & 0.051 & -0.009 & 0.020 & -0.008 \\
Qwen-3-VL-8B-FT & 0.286 & 0.301 & 0.242 & 0.213 & 0.238 & 0.190 & 0.181 & 0.205 & 0.154 \\
VideoPhy2 & - & - & - & - & - & - & 0.302 & 0.303 & 0.247 \\
VMBench & - & - & - & 0.036 & 0.013 & 0.032 & - & - & - \\
WorldModelBench & - & - & - & - & - & - & 0.296 & 0.310 & 0.230 \\
\textbf{Ours} & \textbf{0.761}& \textbf{0.750}& \textbf{0.667}& \textbf{0.633}& \textbf{0.625}& \textbf{0.563}& \textbf{0.727}& \textbf{0.718}& \textbf{0.622}\\
\bottomrule
\end{tabular}
\caption{Correlation with human annotations across three dimensions. Baselines include GPT-5.4~\cite{gpt4o}, Gemini 3.1 Pro~\cite{team2024gemini}, Qwen-3-VL-8B~\cite{bai2025qwen3}, Qwen-3-VL-8B-FT (supervised fine-tuning on \ourdataset-Train), VideoPhy2~\cite{bansal2025videophy}, VMBench~\cite{ling2025vmbench}, and WorldModelBench~\cite{li2025worldmodelbench}.  Independent Human denotes a single volunteer evaluator who did not participate in dataset annotation or calibration, while the benchmark labels are aggregated from three calibrated annotators using MOS.}
\label{tab:main_comparison}

\end{table*}

\begin{table}[t]
\centering
\small
\setlength{\tabcolsep}{4pt}
\renewcommand{\arraystretch}{1.08}
\begin{tabular}{@{}lccc@{}}
\toprule
\textbf{Method} & \textbf{Jaccard} $\uparrow$ & \textbf{Prec.} $\uparrow$ & \textbf{Rec.} $\uparrow$ \\
\midrule
GPT-5.4        & 0.15 & 0.29 & 0.19 \\
Gemini 3.1 Pro & 0.12 & 0.24 & 0.16 \\
Qwen-3-VL-8B   & 0.09 & 0.20 & 0.12 \\
\textbf{Ours}  & \textbf{0.58} & \textbf{0.74} & \textbf{0.63} \\
\bottomrule
\end{tabular}
\caption{
Failure-cause grounding of diagnostic reasoning. We compare GPT-5.4~\cite{gpt4o}, Gemini 3.1 Pro~\cite{team2024gemini}, Qwen-3-VL-8B~\cite{bai2025qwen3}, and our \ourmethod. We parse each model's reasoning into 12 predefined failure causes and compare the parsed causes with human annotations using Jaccard similarity, precision, and recall.
}
\label{tab:rationale_grounding}
\end{table}

\begin{figure}[t]
    \centering
    \includegraphics[width=0.48\textwidth]{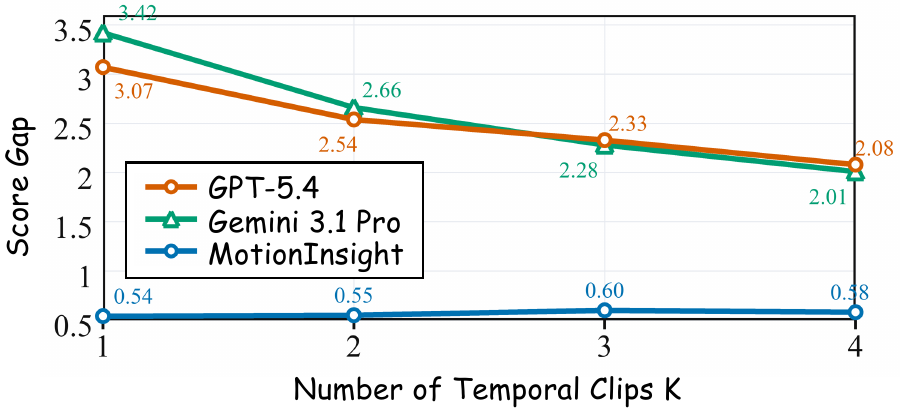}
    \caption{
Sensitivity to localized motion failures. 
We divide each video into $K$ temporal clips and compute the gap between the minimum clip-level score and the human full-video score. 
\ourmethod keeps a consistently small gap across $K$, indicating better sensitivity.
}

    \label{fig:localized-failure-sensitivity}
\end{figure}

\begin{table}[t]
\centering
\small
\setlength{\tabcolsep}{4pt}
\renewcommand{\arraystretch}{1.1}
\begin{tabular}{lcccc}
\toprule
\textbf{Method} & \textbf{\#Fr.} & \textbf{OC}$\uparrow$ & \textbf{MC}$\uparrow$ & \textbf{PP}$\uparrow$ \\
\midrule
Uniform Sampling           & 16 & 0.510 & 0.455 & 0.445 \\
2$\times$ Uniform Sampling & 32 & 0.435 & 0.382 & 0.421 \\
4$\times$ Uniform Sampling & 64 & 0.406 & 0.334 & 0.397 \\
Trajectory Overlay         & 16 & 0.475 & 0.438 & 0.430 \\
Motion-Aware Rep.   & 16 & \textbf{0.761} & \textbf{0.633} & \textbf{0.727} \\
\bottomrule
\end{tabular}
\caption{Ablation on video perception strategies under the same GRPO training, reported with SRCC. Rep. denotes representations. \#Fr. refers to the number of sampled RGB frames, OC: Object Consistency, MC: Motion Continuity, and PP: Physical Plausibility.}
\label{tab:ablation_input}
\end{table}

\begin{table}[t]
\centering
\small
\begin{tabular}{lccc}
\toprule
\textbf{Setting} & \textbf{OC}$\uparrow$ & \textbf{MC}$\uparrow$ & \textbf{PP}$\uparrow$ \\
\midrule
Semantic Alignment Only & 0.253 & 0.183 & 0.224 \\
+ Scoring Reward & 0.703 & 0.588 & 0.670 \\
+ Failure-cause Reward & \textbf{0.761} & \textbf{0.633} & \textbf{0.727} \\
\bottomrule
\end{tabular}
\caption{Ablation on the GRPO design of \ourmethod, reported with SRCC. OC: Object Consistency, MC: Motion Continuity, and PP: Physical Plausibility.}

\label{tab:grpo_ablation}
\end{table}

\myparagraph{Dataset and metrics.}
We use 5K videos from OpenVid~\cite{nan2024openvid} to construct 80,721 QA pairs for semantic alignment. In the GRPO stage, \ourdataset-Train is used to align the evaluator with human judgments, while \ourdataset-Test is used to assess the scoring and diagnostic reasoning of \ourmethod. We use PLCC, SRCC, and KRCC to measure the correlation between predicted scores and human annotations. 
For diagnostic reasoning, we parse each model's reasoning texts into 12 predefined failure causes using GPT-5.1 and evaluate their consistency with annotated failure causes using Jaccard similarity, precision, and recall.

\myparagraph{Baselines.} To evaluate both the scoring and diagnostic reasoning capabilities of \ourmethod, we compare it with several general-purpose VLM evaluators, including GPT-5.4~\cite{gpt4o}, Gemini 3.1 Pro~\cite{team2024gemini}, and Qwen-3-VL-8B~\cite{bai2025qwen3}. 
To ensure a fair comparison, we carefully design and validate a unified prompt for both GRPO training and VLM baselines. 
The prompt explicitly specifies the designated target object and the three evaluation dimensions, requiring each evaluator to reason first and then output continuous scores in a consistent format.
The full prompt template is provided in Appendix~\ref{prompt_template}.
To provide a stronger VLM baseline, we further fine-tune Qwen-3-VL-8B for score regression on \ourdataset-Train using only RGB-frame inputs.
For scoring evaluation, we additionally compare with specialized video evaluators.
For evaluators specifically designed for physical assessment, such as VideoPhy2~\cite{bansal2025videophy} and WorldModelBench~\cite{li2025worldmodelbench}, we compare their scoring results on Physical Plausibility. 
For methods specifically designed for motion assessment, we compare the Motion Smoothness Score in VMBench~\cite{ling2025vmbench} on Motion Continuity. As a reference, we also report the scoring performance of a single volunteer evaluator on \ourdataset-Test.

\subsection{Results}
\myparagraph{Qualitative comparisons to baselines.} For the qualitative results shown in Figure~\ref{fig:qualitative_result}, \ourmethod provides diagnostic reasoning from three perspectives. In the illustrated example, \ourmethod identifies that the foam box becomes shorter after the collision, indicating a structural collapse. Its scores are also better aligned with human judgments, as shown in the accompanying radar chart. More qualitative results are shown in Appendix~\ref{more_qualitative_result_suppl}.

\myparagraph{Scoring alignment with human judgments.}
As shown in Table~\ref{tab:main_comparison}, existing evaluators show limited alignment with human annotations on \ourdataset-Test, highlighting the difficulty of motion fidelity evaluation for designated targets.
VMBench relies on rule-based filtering over pixel-level motion signals, which limits its generalization across diverse dynamic scenarios.
Although fine-tuning Qwen-3-VL-8B on \ourdataset-Train improves adaptation, it still generalizes poorly to \ourdataset-Test.
Other VLM-based evaluators also perform poorly: RGB-frame inputs provide only partial temporal observations and may miss frames containing key artifacts, while many object motion deficiencies remain subtle without explicit target motion modeling.
By contrast, \ourmethod achieves performance comparable to a human evaluator across the three dimensions. 
We also explore how \ourmethod's diagnostic scores along three dimensions can be used to optimize video generation in Appendix~\ref{rm_application}.

\myparagraph{Grounded diagnostic reasoning.}
As shown in Table~\ref{tab:rationale_grounding}, existing VLMs exhibit poor diagnostic reasoning capability. 
Their explanations frequently miss the fine-grained motion deficiencies identified by human annotators, resulting in low consistency with the annotated failure causes. 
In contrast, \ourmethod produces more grounded explanations, achieving higher Jaccard similarity, precision, and recall. 
This suggests that explicitly modeling target motion and incorporating failure-cause supervision help \ourmethod provide more insight into object motion deficiencies.

\myparagraph{Sensitivity to localized motion failures.} 
\label{sensitivity}
To evaluate sensitivity to localized motion failures, we select 327 videos containing object consistency deficiencies from VidMotion-Test for scoring evaluation. 
These deficiencies are localized to the object and often appear only in a few frames, making them a suitable testbed for localized failure detection.
For each video, we divide the full sequence into $K$ temporal clips and compare the minimum clip-level score with the human score assigned to the full video.
If an evaluator performs better when the video is divided into more clips, it suggests that the evaluator cannot reliably detect localized motion failures when the entire video is provided as input ($K=1$), where local artifacts can be diluted by redundant visual tokens.

As shown in Figure~\ref{fig:localized-failure-sensitivity}, GPT-5.4 and Gemini 3.1 Pro exhibit a large score gap when evaluating the full video directly, but the gap decreases as the video is divided into more temporal clips.
In contrast, \ourmethod maintains a consistently small score gap across different values of $K$.
This suggests that \ourmethod is more sensitive to localized motion failures, consistent with the human visual system~\cite{human_visual}.

\subsection{Ablation Study}

\myparagraph{Motion-aware representations outperform RGB-frame sampling.} 
As shown in Table~\ref{tab:ablation_input}, we compare different video input strategies while keeping the GRPO training procedure unchanged.
Specifically, we evaluate standard uniform sampling, denser variants with 2$\times$ and 4$\times$ more sampled frames, as well as a trajectory-visualized setting, where the target object's trajectory is drawn on the sampled frames as visual guidance. The results show that neither increasing the sampling density nor overlaying trajectories leads to clear performance gains.
Moreover, denser sampling introduces a higher computational cost while making it harder for the evaluator to focus on localized artifacts, as the critical motion deficiencies can be diluted by redundant visual tokens. In contrast, our motion-aware representations consistently achieve the best performance across all three evaluation dimensions, suggesting that structuring target motion in motion space helps the evaluator perceive artifacts that may remain subtle in RGB-frame observations. Additional ablations on the design of motion-aware representations are provided in Appendix~\ref{ablation_motion_aware_representation}.

\myparagraph{Motion-specific rewards strengthen the diagnostic ability.} In Table~\ref{tab:grpo_ablation}, we further ablate the reward design used in GRPO in terms of scoring performance. After semantic alignment, the model exhibits limited scoring capability without additional preference alignment. Incorporating the scoring reward substantially improves its performance, while further adding the failure-cause reward provides more explicit supervision on motion failure causes and strengthens diagnostic reasoning, leading to further gains in human-aligned scoring.

\section{Conclusion}
We formulate the diagnostic task of object-centric motion fidelity assessment, evaluated along object consistency, motion continuity, and physical plausibility. To support this study, we introduce \ourdataset, a diagnostic dataset with multi-dimensional scores and failure causes. Built on \ourdataset, we further propose \ourmethod, a diagnostic evaluator that shifts the assessment from implicit RGB-frame observations to explicit motion-space diagnosis. We demonstrate the superior performance of \ourmethod in scoring and interpretable failure diagnosis.

\section*{Limitations}

Although VidMotion is designed for object-centric motion fidelity assessment, its scope is mainly limited to entities with clear spatial boundaries, persistent identities, and trackable motion trajectories. Therefore, VidMotion and MotionInsight may be less suitable for dynamic phenomena such as fluids, smoke, fire, splashes, or highly deformable materials, where object-centric scoring and failure causes can become inherently ambiguous.

In addition, using MotionInsight as a reward model may introduce Goodhart's-law risks. Since the evaluator remains fixed during generator optimization and does not automatically evolve with advancing generators, generators may overfit to its scoring patterns and obtain higher rewards without corresponding improvements in real object motion fidelity.

\section*{Ethical Considerations}

This work complies with the ACL Ethics Policy. \ourdataset is constructed from open-source datasets and curated video sources.
All external datasets, models, and tools used in this work are properly cited and used in accordance with their respective licenses and terms of use.
Our use of these artifacts is limited to academic research on video generation evaluation, which is consistent with their intended research use, where specified.
The artifacts released by this work are intended solely for research purposes.
All collected videos are manually screened before annotation to filter out invalid, sensitive, offensive, or privacy-risk content to the best of our ability, including videos containing personal information, inappropriate content, or clearly privacy-sensitive visual content. 
The released annotations focus only on target objects, motion quality scores, confidence levels, and failure causes, and do not include annotator identities or personally identifying information.

\section*{Acknowledgments}
The work is supported by the National Key R\&D Program of China (No. 2025YFE0201300).

\bibliography{custom}
\clearpage
\appendix

\section{Annotation Details}
\label{data_annotation}

Annotators were recruited from students and researchers with experience in video understanding. All 21 annotators had at least a bachelor's degree. All annotators were fairly compensated for their work. Before participation, annotators were informed that their annotations would be used for academic research on motion fidelity assessment, and they provided consent to participate. 

\begin{figure*}[t]
    \centering
    \includegraphics[width=\textwidth]{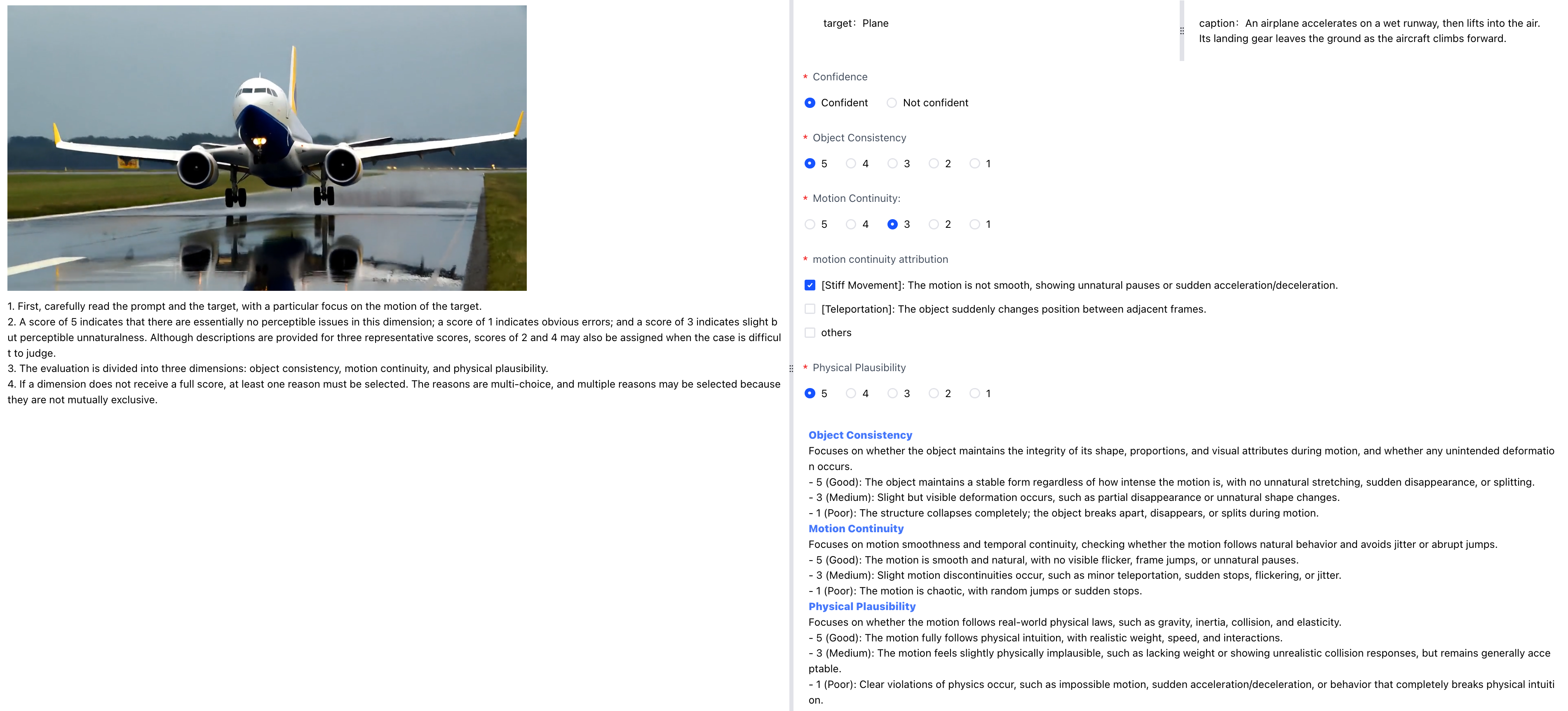}
    \caption{Interface of the annotation website.}
    \label{fig:annotation-web}
\end{figure*}

Figure~\ref{fig:annotation-web} shows the interface of the annotation website. As shown in Table~\ref{tab:annotation_reliability}, Krippendorff's $\alpha$ reaches 0.7981, 0.6846, and 0.7432 for object consistency, motion continuity, and physical plausibility, respectively, indicating reliable inter-annotator agreement across all three scalar dimensions. For the multi-label failure-cause annotations, we compute pairwise Jaccard similarity among annotators by treating each annotator's selected causes as a set. The resulting Jaccard similarity reaches 0.61, suggesting reasonable agreement despite the inherent ambiguity of fine-grained motion artifacts.

\section{\ourdataset Analysis}
\label{dataset_analysis}

\begin{figure}[t]
    \centering
    \includegraphics[width=\linewidth]{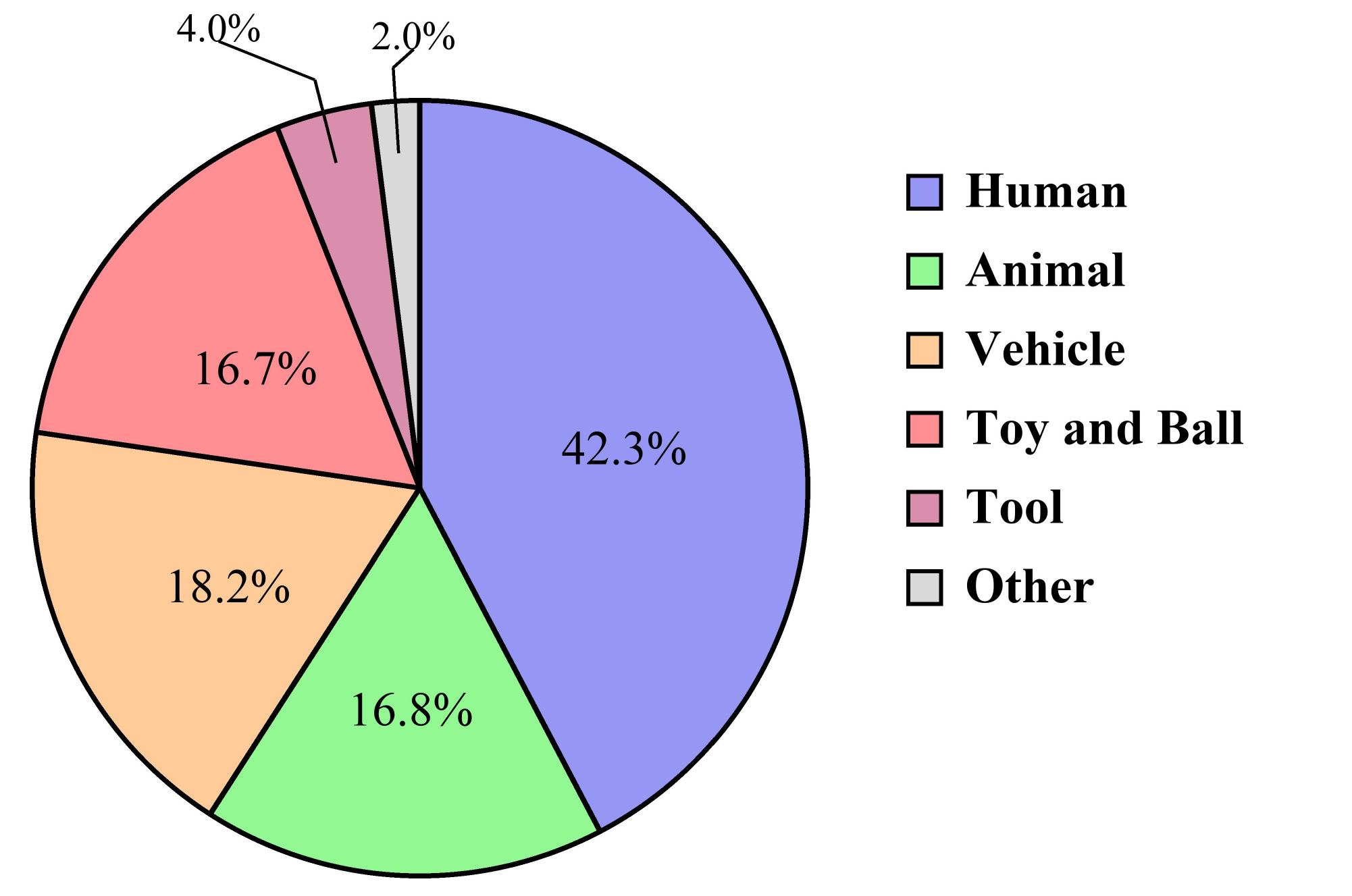}
    \caption{
    Distribution of target objects in \ourdataset. The objects are grouped into six coarse categories: Human, Animal, Vehicle, Toy and Ball, Tool, and Other. The distribution highlights the diversity of object-centric motion scenarios covered by the dataset.
    }
    \label{fig:object-diversity}
\end{figure}

\begin{figure*}[t]
    \centering
    \includegraphics[width=\textwidth]{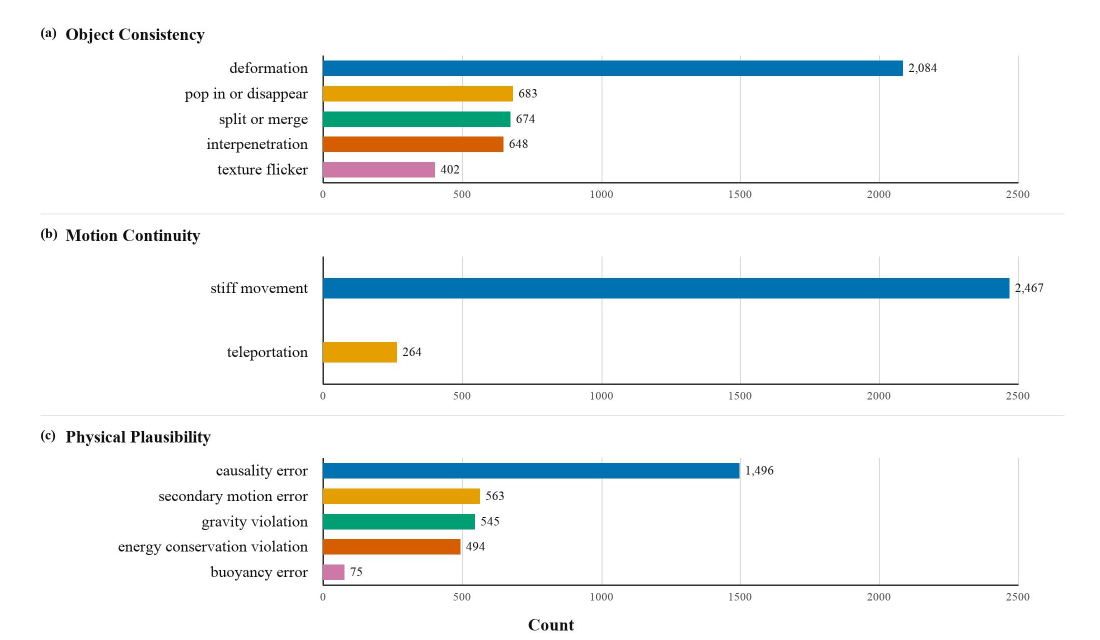}
    \caption{
    Distribution of annotated issue categories. The 12 cause candidates are grouped into three dimensions: Object Consistency, Motion Continuity, and Physical Plausibility.
    }
    \label{fig:error-category-distribution}
\end{figure*}

As shown in Figure~\ref{fig:object-diversity}, \ourdataset shows great diversity in target objects, covering most dynamic objects commonly encountered in daily life. We also analyze the failure cause annotations in Figure~\ref{fig:error-category-distribution}. For object consistency, we find that deformation of humans or objects caused by fast motion is the most common issue. For motion continuity, generated object motions are still often stiff and unnatural. For physical plausibility, current video generation models still struggle with physical causality, frequently producing misaligned motion for observed force.

\section{Prompt Template}
\label{prompt_template}
We present the prompt used for GRPO training and VLM baselines in Table~\ref{tab:score_prompt}.

\section{More Qualitative Results}
\label{more_qualitative_result_suppl}
\begin{figure*}[t]
    \centering
    \includegraphics[width=\linewidth]{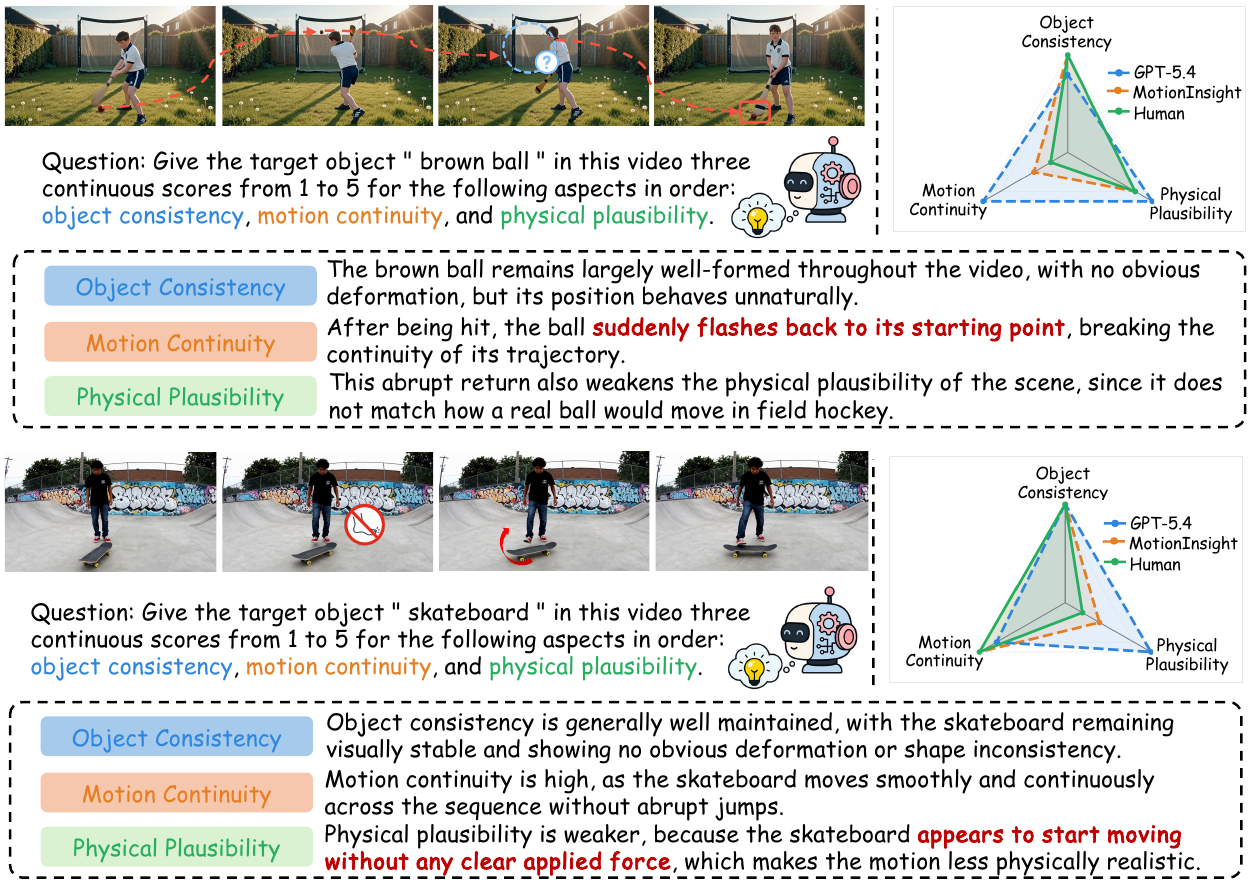}
    \caption{
    Additional qualitative results of \ourmethod. For each example, we show the sampled video frames, the model's reasoning for the designated object, and the predicted scores across the three motion-quality dimensions.
    }
    \label{fig:more_qualitative_result}
\end{figure*}

Figure~\ref{fig:more_qualitative_result} shows two additional examples. In the field hockey case, the ball suddenly flashes back after being hit, which reflects a localized continuity failure. In the skateboard case, the object remains structurally stable but starts moving without a clear applied force, revealing failure in physical plausibility. These results suggest that \ourmethod exhibits reasoning ability for motion, enabling it to evaluate object motion progressively from appearance stability to temporal continuity and even physical fidelity.

\section{Implementation Details} 
We use Qwen-3-VL-8B-Instruct~\cite{bai2025qwen3} as the pretrained VLM backbone. For modality alignment, we fine-tune the motion encoding components for 3 epochs on 8 NVIDIA A100 GPUs with a learning rate of $1\times10^{-6}$. For GRPO training, we set the number of sampled responses to $N=8$ and the KL penalty weight to $\beta=0.001$. The model is then trained for 5 epochs on 8 NVIDIA A100 GPUs with a learning rate of $1\times10^{-6}$. 
In the motion-aware representations, the number of learnable queries $K$ is set to 8.
For the multi-dimensional scoring reward, the asymmetric penalty coefficient $\alpha$ is set to 1.5.
For RGB-frame inputs, \ourmethod samples frames at a stride of 16 frames.

\section{\ourmethod for Generation}
\label{rm_application}
We further explore the potential of \ourmethod as a reward for improving video generation through preference optimization.
We use Wan2.1-T2V-1.3B~\cite{Wan} as the base generator.
For each prompt, we sample eight candidate videos and use Qwen-3-VL-8B-Instruct~\cite{bai2025qwen3} to identify salient moving objects in each video.
For each detected object $o_i$, \ourmethod predicts three scores $s_i^{OC}$, $s_i^{MC}$, and $s_i^{PP}$.
We compute the object-level motion reward by multiplying the normalized scores across the three dimensions:
\begin{equation}
r_i = \prod_{d \in \{OC, MC, PP\}} \frac{s_i^d}{5}.
\end{equation}
We then obtain the video-level object-motion reward by taking the minimum reward over all detected salient moving objects:
\begin{equation}
R(v) = \min_{i=1}^{N} r_i.
\end{equation}
For each prompt, we select the candidate with the highest $R(v)$ as the positive sample and the candidate with the lowest $R(v)$ as the negative sample, forming a preference pair for DPO training.
We then fine-tune the base generator using these \ourmethod-derived preference pairs.
For comparison, we construct another set of preference pairs using VideoPhy2~\cite{bansal2025videophy} scores under the same candidate pool and train a corresponding DPO baseline.

\begin{figure*}[t]
    \centering
    \includegraphics[width=\textwidth]{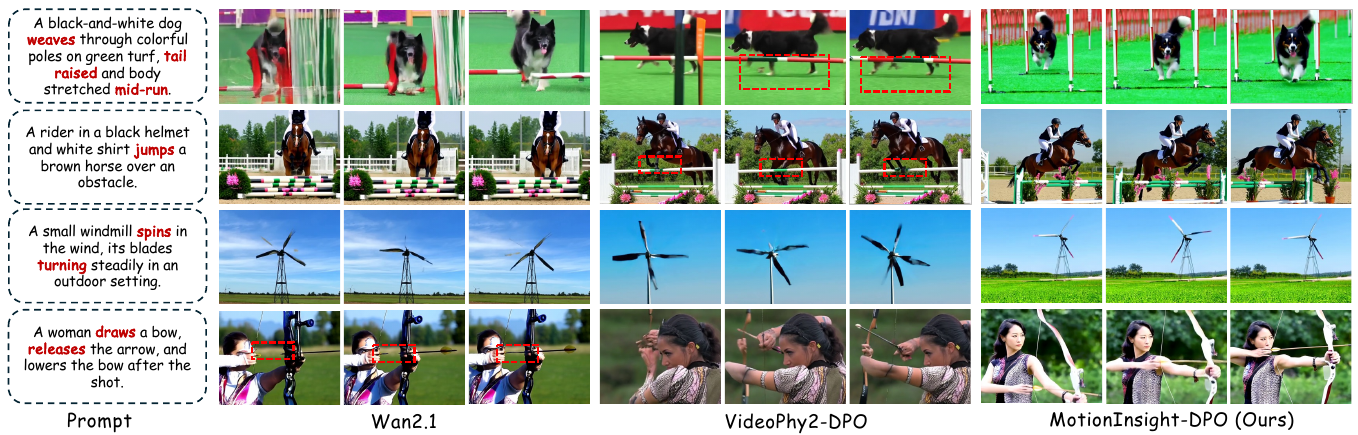}
    \caption{
    Qualitative comparison of DPO-based video generation. 
    }
    \label{fig:dpo_qualitative}
\end{figure*}

We recruit 20 volunteers to conduct a 2AFC human study.
As shown in Table~\ref{tab:2AFC}, DPO training with \ourmethod-derived preference pairs is strongly preferred over both the original Wan 2.1 model and the VideoPhy2-DPO baseline.
Figure~\ref{fig:dpo_qualitative} further shows that \ourmethod-DPO reduces typical object motion artifacts, including deformation during bow drawing and foot penetration when the horse crosses a hurdle.
These results demonstrate that the diagnostic signals from \ourmethod are useful not only for evaluation, but also for improving video generation, leading to better object motion fidelity while preserving overall video quality.

\section{Ablation on Motion-Aware Representations}
\label{ablation_motion_aware_representation}
Table~\ref{tab:ablation_embedding} further validates the effectiveness of the proposed motion-aware representation. When the temporal order of the motion embeddings is randomly shuffled, performance drops substantially across all three dimensions, indicating that the model relies on temporally structured motion form rather than merely benefiting from additional features. Moreover, retaining only object-motion tokens also underperforms the full model, showing that both object motion and camera poses are important for disentangling object dynamics from camera-induced motion.

\begin{table}[t]
\centering
\small
\setlength{\tabcolsep}{4pt}
\renewcommand{\arraystretch}{1.1}
\begin{tabular}{lccc}
\toprule
\textbf{Method} & \textbf{OC}$\uparrow$ & \textbf{MC}$\uparrow$ & \textbf{PP}$\uparrow$ \\
\midrule
Motion-Aware Rep.        & \textbf{0.761} & \textbf{0.633} & \textbf{0.727} \\
w/ Temporal Shuffle      & 0.161 & 0.109 & 0.113 \\
w/o Camera Motion        & 0.697 & 0.580 & 0.683 \\
\bottomrule
\end{tabular}
\caption{Ablation on the design of motion-aware representations for GRPO, reported with SRCC. Rep. denotes representations. w/ Temporal Shuffle randomly permutes the temporal order of the motion representations, while w/o Camera Motion removes camera-motion tokens and keeps only object-motion features. OC: Object Consistency, MC: Motion Continuity, and PP: Physical Plausibility.}
\label{tab:ablation_embedding}
\end{table}

\begin{table}[t]
\centering
\small
\begin{tabular}{lc}
\toprule
Dimension & Krippendorff's $\alpha$ \\
\midrule
Object Consistency & 0.7981 \\
Motion Continuity & 0.6846 \\
Physical Plausibility & 0.7432 \\
\bottomrule
\end{tabular}
\caption{
Inter-annotator reliability for the three scalar motion-quality dimensions in \ourdataset. Krippendorff's $\alpha$ is used to measure agreement among annotators, with higher values indicating stronger consistency.
}
\label{tab:annotation_reliability}
\end{table}

\begin{table}[h]
\vspace{.5em}
\centering
\small
\setlength{\tabcolsep}{4pt}
\begin{tabular}{lcc}
\toprule
Method & Motion Fidelity & Overall Quality \\
\midrule
 over Wan 2.1 & 92.86\%& 86.67\%\\
 over VideoPhy2-DPO & 87.56\%& 87.50\%\\
\midrule
\end{tabular}
\vspace{-.7em}
\caption{2AFC human study results for DPO-based video generation. 
We compare the model trained with \ourmethod-derived preference pairs against the original Wan2.1 and the VideoPhy2-DPO~\cite{bansal2025videophy} baseline.}
\vspace{-2em}
\label{tab:2AFC}
\end{table}

\begin{table*}[t]
\centering
\begin{minipage}{0.82\textwidth}
\hrule height 1.0pt
\vspace{0.6em}

\small
Suppose you are an expert in judging and evaluating object-centric motion quality in real or AI-generated videos. Given the sampled frames of a video and the target object to be evaluated, please carefully analyze the motion of the target object.\\

You should evaluate the target object's motion from three dimensions:\\
(1) object consistency: whether the target object preserves a stable and coherent appearance, identity, and structure during motion;\\
(2) motion continuity: whether the target object's motion is temporally smooth, continuous, and natural;\\
(3) physical plausibility: whether the target object's motion follows intuitive physical causality and common-sense dynamics.

\vspace{1.2em}

Please first provide your reasoning process within \texttt{<thinking>} \texttt{</thinking>} tags, and then output only the three scores within \texttt{<answer>} \texttt{</answer>} tags. Each score should be in [1, 5], with one decimal place. For each dimension, output a score from [1, 5], where 1 means very poor, 3 means minor but noticeable artifacts, and 5 means excellent. Use the following exact JSON format:\\

\texttt{<thinking>}\\
reasoning process here\\
\texttt{</thinking>}\\
\texttt{<answer>}\\
\{\texttt{"object\_consistency"}: 3.2, \texttt{"motion\_continuity"}: 2.8, \texttt{"physical\_plausibility"}: 3.5\}\\
\texttt{</answer>}

\vspace{1.2em}

Be strict and conservative in scoring. A mostly static object is not necessarily poor if it is consistent with the scene.

\vspace{1.2em}

For this video, the target object is ``\{\textbf{target\_object}\}''.\\
The sampled video frames are as follows:

\vspace{0.6em}
\hrule height 1.0pt
\end{minipage}

\caption{Prompting template used for multi-dimensional scoring reward in \ourmethod GRPO training and VLM baselines.}
\label{tab:score_prompt}
\end{table*}

\begin{figure*}[t]
    \centering
    \includegraphics[width=\linewidth]{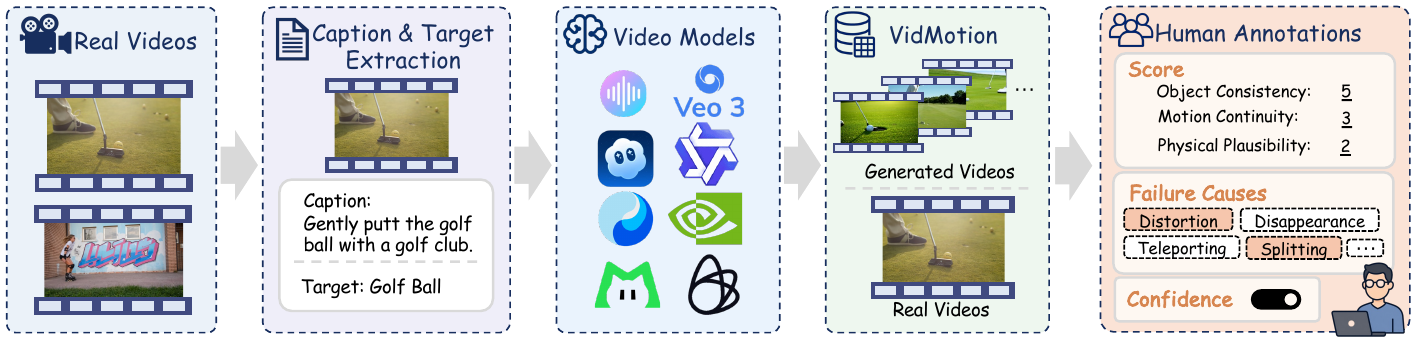}
    \caption{Overview of the \ourdataset construction and annotation pipeline. We collect real videos with prominent object motion, extract captions and target objects, generate corresponding videos using nine open-source and proprietary models, and collect expert annotations for both real and generated videos, including three-dimensional scores, failure causes, and confidence levels.}

    \label{fig:dataset}
\end{figure*}

\end{document}